\documentclass[runningheads]{llncs}
\usepackage{graphicx}
\usepackage{tikz}
\usepackage{comment}
\usepackage{amsmath,amssymb} 
\usepackage{color}
\usepackage{array}
\usepackage{hyperref}
\usepackage{wrapfig}
\usepackage{bm}
\usepackage{multicol,multirow}
\usepackage{algorithm,algorithmic}
\newcommand{\xu}[1]{\textcolor{black}{#1}}
\newcommand{\li}[1]{\textcolor{black}{#1}}
\newcommand{\red}[1]{\textcolor{black}{#1}}
\usepackage{lipsum}
\usepackage{arydshln}

\usepackage[accsupp]{axessibility}
\makeatletter
  \newcommand\tabcaption{\def\@captype{table}\caption}
\makeatother
\usepackage{subfigure}
\usepackage{booktabs}
\usepackage{bbding}
\usepackage{arydshln}
\usepackage[marginal]{footmisc}
\renewcommand{\thefootnote}{}

\begin{document}
\pagestyle{headings}
\mainmatter
\def\ECCVSubNumber{4277}  
\title{IDa-Det: An Information Discrepancy-aware Distillation for 1-bit Detectors}
\titlerunning{IDa-Det} 
\authorrunning{Xu et al.} 
\author{Sheng~Xu\textsuperscript{1}$^{\dagger}$,~Yanjing~Li\textsuperscript{1}$^{\dagger}$, Bohan Zeng\textsuperscript{1}$^{\dagger}$, Teli Ma\textsuperscript{2}, Baochang Zhang\textsuperscript{1,3}$^*$, Xianbin Cao\textsuperscript{1}, Peng Gao\textsuperscript{2}, Jinhu L\"u\textsuperscript{1,3}}
\institute{\textsuperscript{1} Beihang University, Beijing, China\\
\textsuperscript{2} Shanghai Artificial Intelligence Laboratory, Shanghai, China\\
\textsuperscript{3} Zhongguancun Laboratory, Beijing, China\\
\email{\{shengxu, yanjingli, bohanzeng, bczhang\}@buaa.edu.cn}}

\maketitle
\newcommand\blfootnote[1]{%
\begingroup 
\renewcommand\thefootnote{}\footnotetext{#1}%
\addtocounter{footnote}{0}%
\endgroup 
}
\blfootnote{$\dagger$ Equal contribution.}
\blfootnote{$*$ Corresponding author.}
\begin{abstract}
Knowledge distillation (KD) has been proven to be useful for training compact object detection models. However, we observe that KD is often effective when the teacher model and student counterpart share similar proposal \xu{information}. This explains why existing KD methods are less effective for 1-bit detectors, caused by a significant information discrepancy between the real-valued teacher and the 1-bit student. This paper presents an Information Discrepancy-aware strategy (IDa-Det) to distill 1-bit detectors that can effectively eliminate information discrepancies and significantly reduce the performance gap between a 1-bit detector and its real-valued counterpart. \li{We formulate the distillation process as a bi-level optimization formulation. At the inner level, we select the representative proposals with maximum information discrepancy.} We then introduce a novel entropy distillation loss to reduce the disparity based on the selected proposals. Extensive experiments demonstrate IDa-Det's superiority over state-of-the-art 1-bit detectors and KD methods on both PASCAL VOC and COCO datasets. IDa-Det achieves a 76.9\% mAP for a 1-bit Faster-RCNN with ResNet-18 backbone. Our code is open-sourced on \url{https://github.com/SteveTsui/IDa-Det}.
\end{abstract}

\section{Introduction}

Recently, the object detection task \cite{voc2007,coco2014} has been greatly promoted due to advances in deep convolutional neural networks (DNNs) \cite{he2016deep}. However, DNN models comprise a large number of parameters and floating-point operations (FLOPs), restricting their deployment on embedded platforms. Techniques such as compact network design \cite{howard2017mobilenets,ma2018shufflenet}, network pruning \cite{he2018soft,li2016pruning,zhuo2020cogradient}, low-rank decomposition \cite{denil2013predicting}, and quantization \cite{rastegari2016xnor,xu2021poem,zhao2022towards} have been developed to address these restrictions and accomplish an efficient inference on the detection 
task. Among 
\begin{wrapfigure}[23]{r}{0.53\textwidth}
	\centering
	\includegraphics[scale=.15]{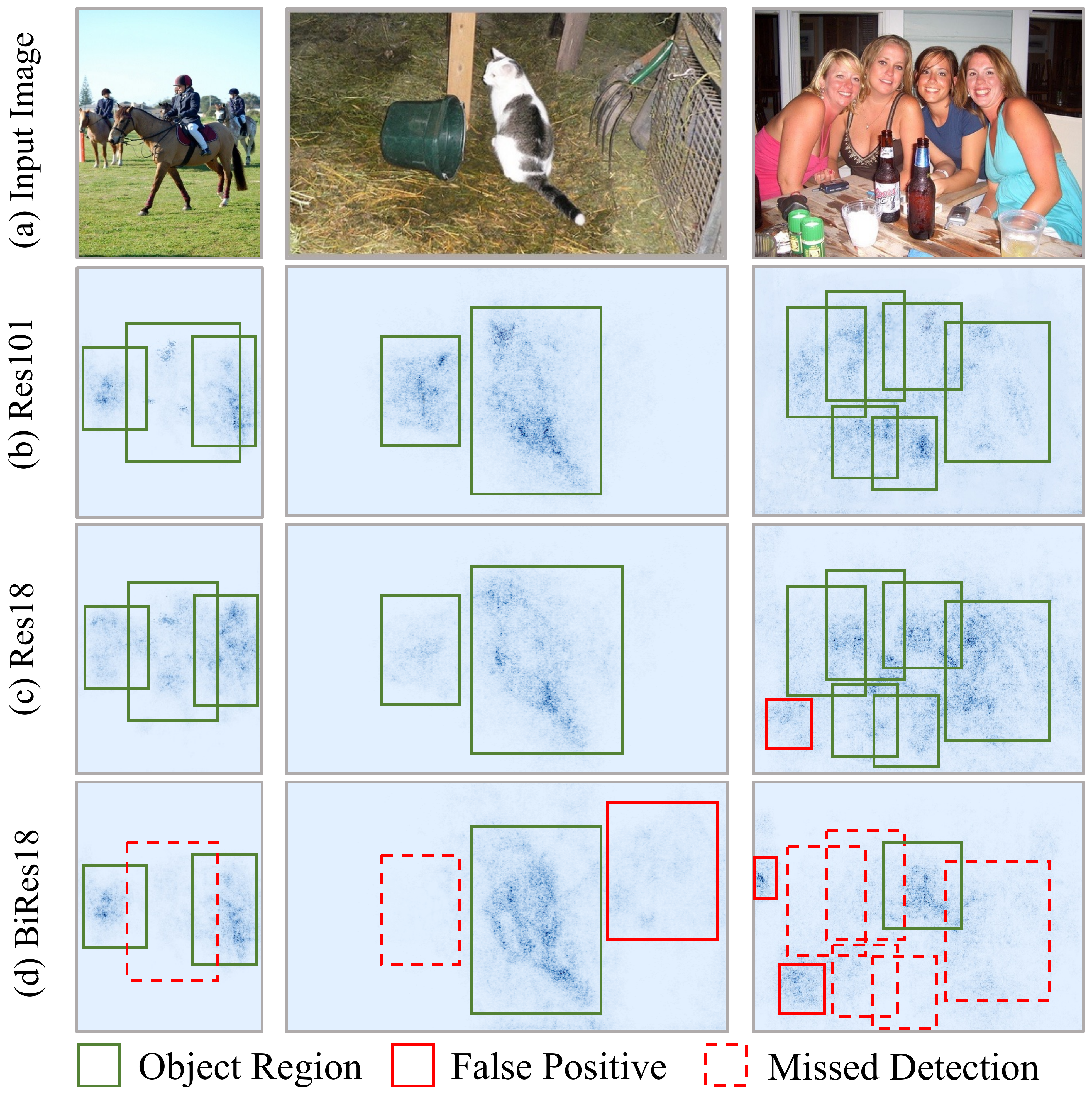}
	\caption{Input images and saliency maps following \cite{guo2021distilling}. Images are randomly selected from VOC {\tt test2007}.  Each row includes: (a) input images, saliency maps of (b) Faster-RCNN with ResNet-101 backbone (Res101), (c) Faster-RCNN with ResNet-18 backbone (Res18), (d) 1-bit Faster-RCNN with ResNet-18 backbone (BiRes18), respectively.} 
	\label{motivation}
\end{wrapfigure}these, binarized detectors have contributed to object detection by accelerating the CNN feature extracting for real-time bounding box localization and foreground classification \cite{xu2020amplitude,wang2020bidet,xu2021layer}. For example, the 1-bit SSD300 \cite{liu2016ssd} with VGG-16 backbone \cite{Simonyan15} theoretically achieve the acceleration rate up to 15$\times$ with XNOR and Bit-count operations using binarized weights and activations as described in \cite{wang2020bidet}. 
With extremely high energy-efficiency for embedded devices, they are able to be installed directly on next-generation AI chips. Despite these appealing features, 1-bit detectors' performance often deteriorates to the point, which explains why they are not widely used in real-world embedded systems.

\begin{figure}[t]
    \begin{minipage}[t]{\linewidth}
	\subfigure[\scriptsize{VOC {\tt trainval0712}}]{
		\begin{minipage}[t]{0.25\textwidth}
			\centering
			\includegraphics[width= \linewidth]{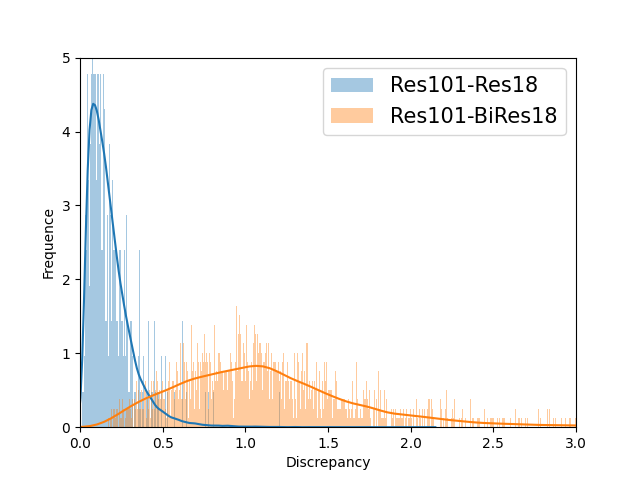}
		\end{minipage}%
	}%
	\subfigure[\scriptsize{VOC {\tt test2007}}]{
		\begin{minipage}[t]{0.25\textwidth}
			\centering
			\includegraphics[width= \linewidth]{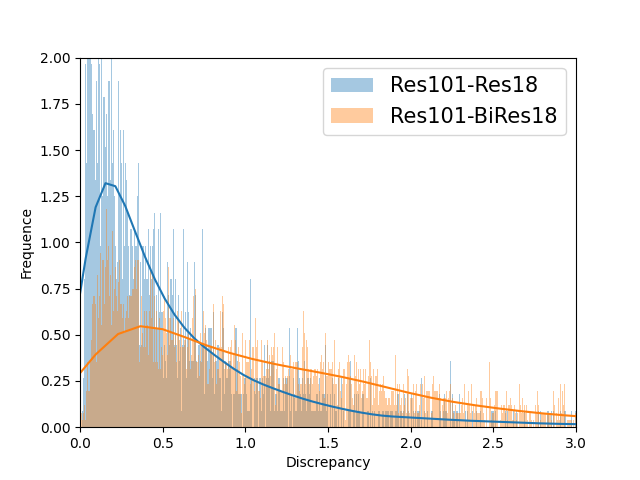}
		\end{minipage}%
	}%
	\subfigure[\scriptsize{COCO {\tt trainval35k}}]{
		\begin{minipage}[t]{0.25\textwidth}
			\centering
			\includegraphics[width= \linewidth]{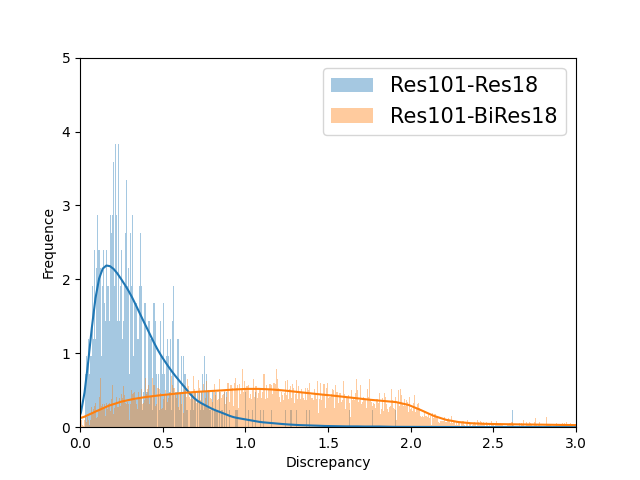}
		\end{minipage}%
	}%
	\subfigure[\scriptsize{COCO {\tt minival}}]{
		\begin{minipage}[t]{0.25\textwidth}
			\centering
			\includegraphics[width= \linewidth]{result_distance_train.png}
		\end{minipage}%
	}%
	\caption{The Mahalanobis distance of the gradient in the intermediate neck feature between Res101-Res18 (blue) and Res101-BiRes18 (orange) in various datasets.}
	\label{dist}
	\end{minipage}
\end{figure}

The recent art \cite{xu2021layer} employs fine-grained feature imitation (FGFI) \cite{wang2019distilling} to enhance the performance of 1-bit detectors. However, it neglects the intrinsic information discrepancy between 1-bit detectors and real-valued detectors. As shown in Fig. \ref{motivation}, we demonstrate that saliency maps of real-valued Faster-RCNN of the ResNet-101 backbone (often used as the teacher network) and the ResNet-18 backbone, in comparison with 1-bit Faster-RCNN of the ResNet-18 backbone (often used as the student network) from top to bottom. \xu{They show that knowledge distillation (KD) methods like \cite{wang2019distilling} are effective for distilling real-valued Faster-RCNNs, only when their teacher model and their student counterpart \li{share small information discrepancy} on proposals, as shown in Fig. \ref{motivation} (b) and (c). This phenomenon does not happen for 1-bit Faster-RCNN, as shown in Fig. \ref{motivation} (b) and (d).  This might explain why existing KD methods are less effective in 1-bit detectors. \li{A statistic on COCO and PASCAL VOC datasets in Fig. \ref{dist} show that  the discrepancy between proposal saliency maps of Res101 and Res18 (blue) is much smaller than that of Res101 and BiRes18 (orange). That is to say, the smaller the distance is, the less the discrepancy is.} Briefly, conventional KD methods show their effectiveness on distilling real-valued detectors but seem to be less \xu{effective} on distilling 1-bit detectors.}

\begin{figure}[t]
	\centering
	\includegraphics[scale=.3]{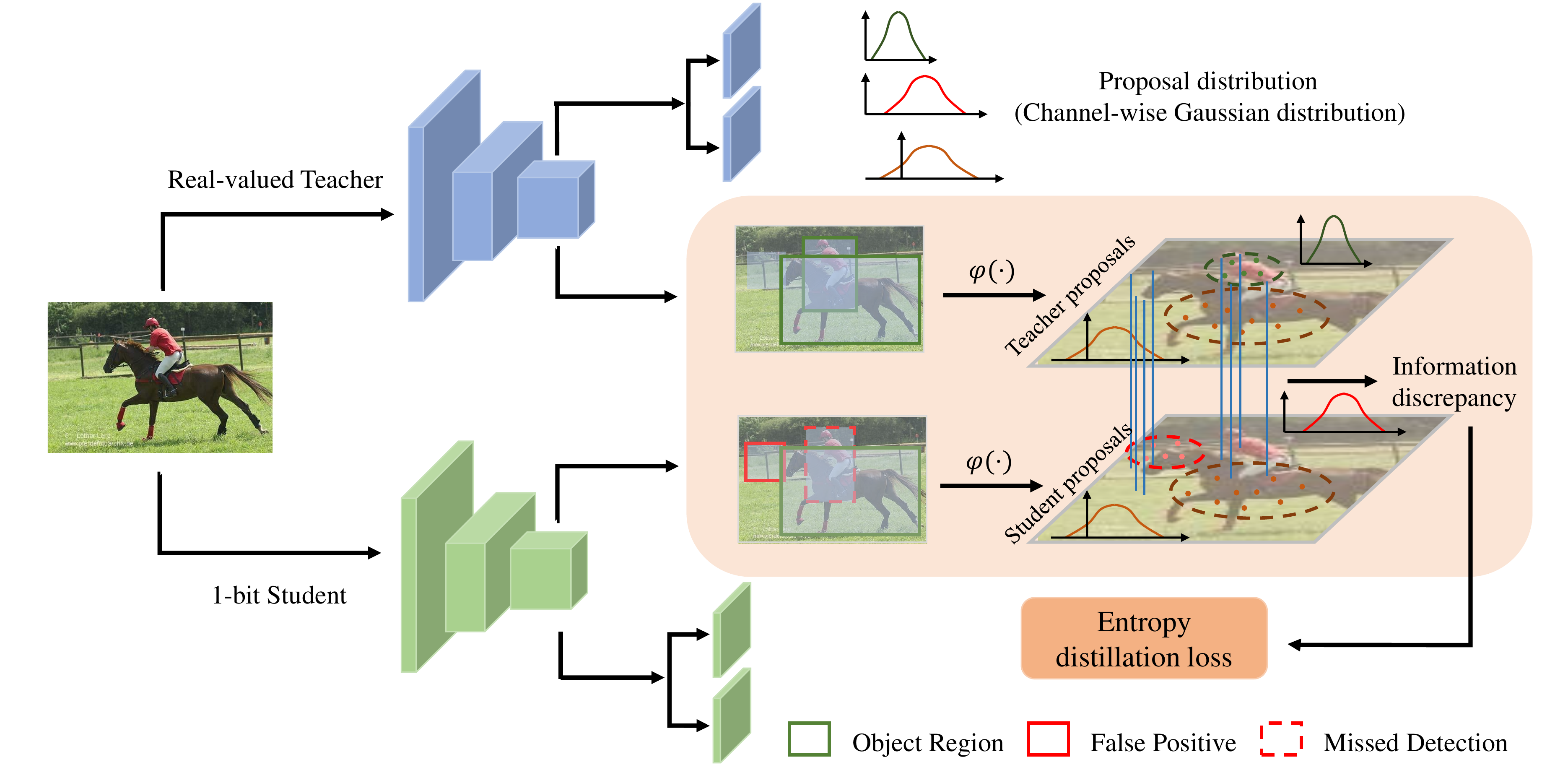}\\
	\caption{Overview of the proposed information discrepancy-aware distillation (IDa-Det) framework. We \li{first select representative proposal pairs based on the information discrepancy. Then we propose the entropy distillation loss to eliminate the information discrepancy.}}
	\label{framework}
\end{figure}

In this paper, we are motivated by the above observation and present an information discrepancy-aware distillation for 1-bit detectors (IDa-Det), which can effectively address the information discrepancy problem, leading to an efficient distillation process. As shown in Fig. \ref{framework}, \li{we introduce a discrepancy-aware method to select proposal pairs and facilitate distilling 1-bit detectors, rather than only using object anchor locations of student models or ground truth as in existing methods  \cite{wang2019distilling,xu2021layer,guo2021distilling}.}
We further introduce a novel entropy distillation loss to \xu{leverage} more comprehensive information than the conventional loss functions. By doing so, we achieve a powerful information discrepancy-aware distillation method for 1-bit detectors (IDa-Det). Our contributions are summarized as: 
\begin{itemize}
	\item \li{Unlike existing KD methods, we distill 1-bit detectors by fully considering the information discrepancy into optimization, which is simple yet effective for learning 1-bit detectors.}
	\item We propose an entropy distillation loss further to improve the representation ability of the 1-bit detector and effectively eliminate the information discrepancy.
	\item \red{We compare our IDa-Det against state-of-the-art 1-bit detectors and KD methods on the VOC and large-scale COCO datasets. Extensive results reveal that our method outperformas the others by a considerable margin. For instance, on VOC {\tt test2007}, the 1-bit Faster-RCNN with ResNet-18 backbone achieved by IDa-Det obtains 76.9\% mAP, achieving a new state-of-the-art.}
 \end{itemize}

\section{Related Work}
\noindent\textbf{1-bit Detectors.}
\red{By removing the foreground redundancy, BiDet \cite{wang2020bidet} fully exploits the representational capability of the binarized convolutions. In this way, the information bottleneck is introduced, which limits the amount of data in high-level feature maps, while maximizing the mutual information between feature maps and object detection. 
The performance of the Faster R-CNN detector is significantly enhanced by the ASDA-FRCNN \cite{xu2020amplitude} which suppresses the shared amplitude between the real-value and binary kernels.
LWS-Det \cite{xu2021layer} novelly proposes a layer-wise searching approach, minimizing the angular and amplitude errors for 1-bit detectors. Also, FGFI \cite{wang2019distilling} is used by LWS-Det to distill the backbone feature map further.}

\noindent\textbf{Knowledge Distillation.}
Knowledge distillation (KD), a significant subset of model compression methods, aims to transfer knowledge from a well-trained teacher network to a more compact student model.
The student is supervised using soft labels created by the teacher, as firstly proposed by \cite{jimmy2014do}. Knowledge distillation is redefined by \cite{hinton2014distilling} as training a shallower network after the softmax layer to approximate the teacher's output. Object detectors can be compressed using knowledge distillation, according to numerous recent papers. 
Chen {\em et al.} \cite{chen2017learning} distill the student through all backbone features, regression head, and classification head, but both the imitation of whole feature maps and the distillation in classification head fail to add attention to the important foreground, potentially resulting in a sub-optimal result. 
Mimicking \cite{li2017mimicking} distills the features from sampled region proposals. However, just replicating the aforementioned regions may lead to misdirection, because the proposals occasionally perform poorly. 
In order to distill the student, FGFI \cite{wang2019distilling} introduces a unique attention mask to create fine-grained features from foreground object areas. DeFeat \cite{guo2021distilling} balances the background and foreground object regions to efficiently distill the student.

In summary, existing KD frameworks for object detection can only be employed for real-valued students having similar information as their teachers. Thus, they are often ineffective in distilling 1-bit detectors. Unlike prior arts, we identify that the information discrepancy between real-valued teacher and 1-bit students are significant for distillation. We first introduce Mahalanobis distance to \xu{identify} the information discrepancy and then accordingly distill the features. Meanwhile, we propose a \xu{novel entropy distillation loss} to prompt the discrimination ability of 1-bit detectors further.

\section{The Proposed Method}
\label{sec4}
In this section, we describe our IDa-Det in detail. Firstly, we overview the 1-bit CNNs. We then describe how we employ the information discrepancy method (IDa) to select representative proposals. Finally, we describe the entropy distillation loss to delicately eliminate the information discrepancy between the real-valued teachers and the 1-bit students. 

\subsection{Preliminaries}
In a specific convolution layer, ${\bf w}\in \mathbb{R}^{C_{out}\times C_{in}\times K \times K}$, ${\bf a}_{in}\in \mathbb{R}^{C_{in} \times W_{in} \times H_{in}}$, and ${\bf a}_{out}\in \mathbb{R}^{C_{out} \times W_{out} \times H_{out}}$ represent its weights and feature maps, where  $C_{in}$ and $C_{out}$ represents the number of channels. $(H, W)$ are the height and width of the feature maps, and $K$ denotes the kernel size. We then have
\begin{equation}
\small
{\bf a}_{out} = {\bf a}_{in} \otimes {\bf w},
\end{equation}
where $\otimes$ is the convolution operation. We omit the batch normalization (BN) and activation layers for simplicity. The 1-bit model aims to quantize ${\bf w}$ and ${\bf a}_{in}$ into ${\bf b}^{\bf w}\in \{-1,+1\}^{C_{out}\times C_{in} \times K \times K}$ and ${\bf b}^{{\bf a}_{in}}\in \{-1,+1\}^{C_{in} \times H \times W}$ using the efficient XNOR and Bit-count  operations to replace full-precision operations. Following \cite{courbariaux2015binaryconnect,courbariaux2016binarized}, the forward process of the 1-bit CNN is
\begin{equation}
\small
{{\bf a}_{out}} = \bm{\alpha} \circ {{\bf b}^{{\bf a}_{in}}} \odot {{\bf b}^{\bf w}},
\label{2}
\end{equation}
where $\odot$ is the XNOR, and bit-count operations and $\circ$ denotes the channel-wise multiplication. $\bm{\alpha}=[\alpha_1,\cdots,\alpha_{C_{out}}]\in \mathbb{R}_+$ is the vector consisting of channel-wise scale factors. ${\bf b} = {\rm sign(\cdot)}$ denotes the binarized variable using the sign function, which returns 1 if the input is greater than zero, and -1 otherwise. It then enters several non-linear layers, {\em e.g.}, BN layer, non-linear activation layer, and max-pooling layer. We omit these for simplification. And then, the output ${\bf a}_{out}$ is binarized to ${\bf b}^{{\bf a}_{out}}$ via the sign function. The fundamental objective of BNNs is calculating ${\bf w}$. We want it to be as close as possible before and after binarization, such that the binarization effect is minimized. Then, we define the reconstruction error as 
\begin{equation}
\small
{L}_R({\bf w},\bm{\alpha}) = {\bf w}-\bm{\alpha}\circ {\bf b}^{\bf w}.
\label{4}
\end{equation}

\subsection{Select proposals with Information Discrepancy}

To eliminate the large  magnitude scale difference   between the real-valued teacher and the 1-bit student, we introduce a channel-wise transformation for the proposals\footnote{In this paper, the proposal denotes the neck/backbone feature map patched by the region proposal of detectors.} of the intermediate neck. We first apply a transformation $\varphi(\cdot)$ on a proposal $\tilde{R}_n\in\mathbb{R}^{C\times W\times H}$ and have 
\begin{equation}
    {R}_{n;c}(x,y)= \varphi(\tilde{R}_{n;c}(x,y)) = \frac{\operatorname{exp}(\frac{\tilde{R}_{n;c}(x,y)}{\mathcal{T}})}{\sum_{(x',y') \in (W, H)}\operatorname{exp}(\frac{\tilde{R}_{n;c}(x'y')}{\mathcal{T}})},
\end{equation}
\begin{wrapfigure}{r}{0.4\textwidth}
	\centering
	\includegraphics[scale=.28]{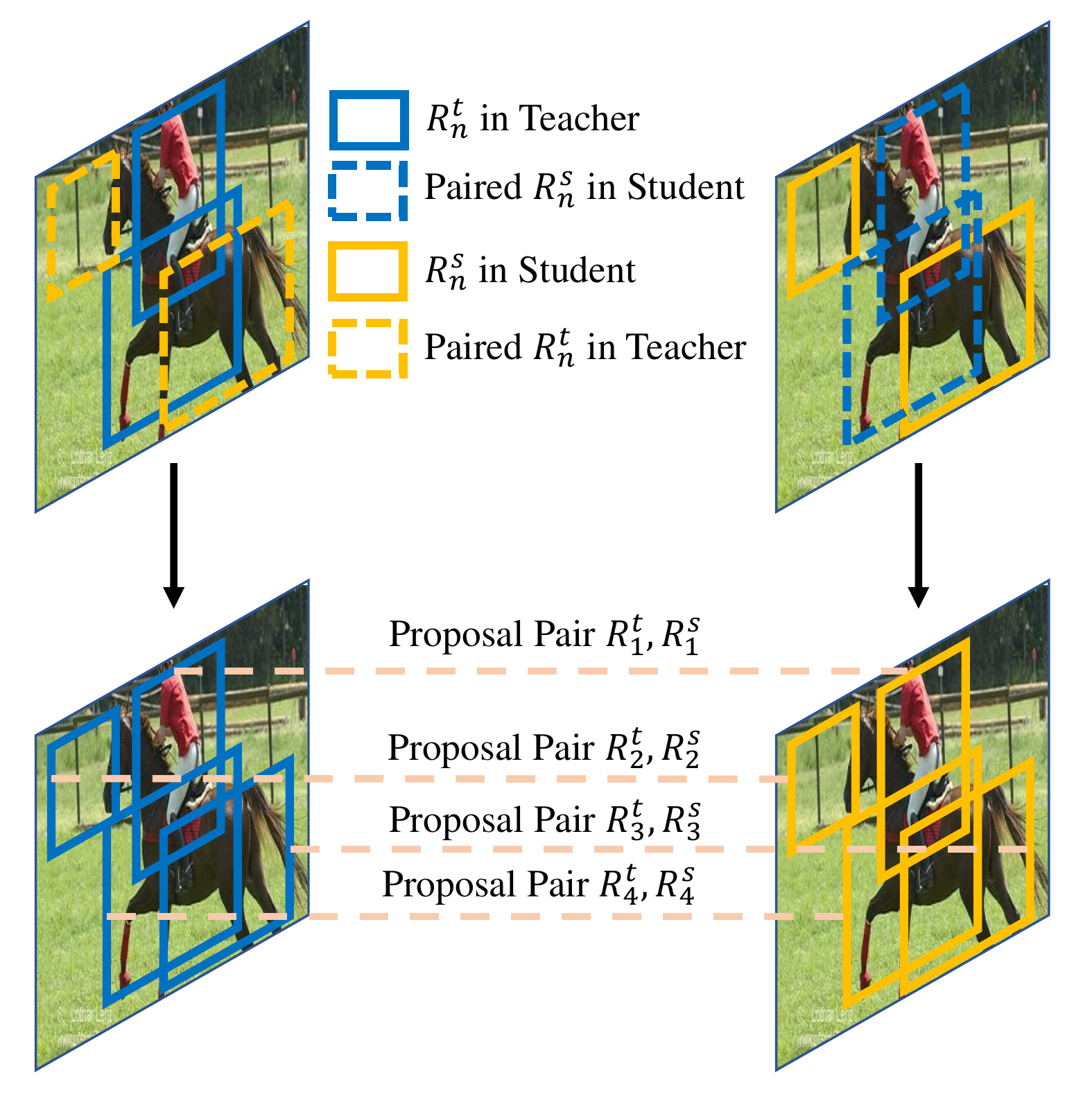}
	\caption{Illustration for the generation of the proposal pairs. Each single proposal in one model generates a counterpart feature map patch in the same location of the other model.}
	\label{proposal}
\end{wrapfigure}
where $(x,y)\in(W, H)$ denote a specific spatial location $(x,y)$ in spatial range $(W, H)$, and $c \in \{1, \cdots, C\}$ is the channel index. $n\in\{1,\cdots,N\}$ is the  proposal index. $N$ denotes the number of proposals. $\mathcal{T}$ denotes a hyper-parameter controlling the statistical attributions of the channel-wise alignment operation\footnote{In this paper, we set $\mathcal{T}=4$.}. 
After the transformation, features in each channel of a proposal projected into the same feature space \cite{wang2021distilling} and follow a Gaussian distribution as 
\begin{equation}
    {p}({R}_{n;c}) \sim \mathcal{N}(\mu_{n;c}, {\sigma^2_{n;c}}).
    \label{eq_distribution}
\end{equation}
We further evaluate the information discrepancy between the proposals of the teacher and the student. As shown in Fig. \ref{proposal}, the teacher and student have $N_T$ and $N_S$ proposals, respectively. Each single proposal in one model generates a counterpart feature map patch in the same location of the other model, thus total $N_T+N_S$ proposal pairs are considered. To evaluate the information discrepancy, we introduce the Mahalanobis distance of each channel-wise proposal feature and measure the discrepancy as
\begin{equation}
\begin{aligned}
  {\varepsilon}_n = \sum_{c=1}^{C}||({R}^t_{n;c} - {R}^s_{n;c})^T{\Sigma^{-1}_{n;c}}
  ({R}^t_{n;c} - {R}^s_{n;c})||_2,
  \label{Mahalanobis}
\end{aligned}
\end{equation}
where $\Sigma_{n;c}$ denotes the covariance matrix of the teacher and student in the $c$-th channel of the $n$-th proposal pair.
\li{The Mahalanobis distance takes both the pixel-level distance between proposals and the statistical characteristics differences in proposal pairs into account.}

\li{To select representative proposals with maximum information discrepancy}, we first define a binary distillation mask $m_n$ as
\begin{equation}
\small
  \begin{aligned}
  m_n = \left\{
  \begin{aligned}
  &1, \;\;\text{if pair $(R_n^t, R_n^s)$ is selected} \\
  &0, \;\;\text{otherwise}
  \end{aligned}
  \right. 
  \end{aligned}
  \label{m}
\end{equation}
where $m_n=1$ denotes distillation will be applied on such proposal pair, otherwise remain unchanged. For each proposal pair, only when their distribution is quite different,  the student model can learn from the teacher counterpart, where a distillation process is needed. 

Based on the derivation above, discrepant proposal pairs will be optimized through distillation. For distilling the selected pairs, we resort to maximize conditional probability $p({R}_n^s |{R}_n^t)$. That is to say, after distillation or optimization, feature distributions of teacher proposal and student counterpart become similar with each other.
To this end, we define $p({R}_n^s |{R}_n^t)$ with $m_n, n\in\{1, \cdots, N_T+N_S\}$ in consideration as 
\begin{equation}
\small
p({R}_n^s |{R}_n^t; m_n) \sim {m_n}\mathcal{N}(\mu_n^t, {\sigma_n^t}^2) + (1 - m_n)\mathcal{N}(\mu_n^s, {\sigma_n^s}^2).
\label{distribution}
\end{equation}
\li{Subsequently, we introduce a bi-level optimization formulation to solve the distillation problem as}
\begin{equation}
\begin{aligned}
\mathop{\max}_{{R}_{n}^s} \; &p({R}_{n}^s|{R}_{n}^t; {\bf m^{*}}), \;\; \forall\;n \in \{0, \cdots, N_T+N_S\},\\
{\rm s.t.}\;\;&{\bf m^{*}} = \mathop{\arg\max}_{{\bf m}}\sum^{N_T+N_S}_{n=1} m_n\varepsilon_n,
\label{objective}
\end{aligned}
\end{equation}
where ${\bf m}=[m_1,\cdots,m_{N_T+N_S}]$ and $||{\bf m}||_0=\gamma\cdot(N_T+N_S)$. $\gamma$ is a hyper-parameter. 
In this way, we select $\gamma\cdot(N_T+N_S)$ pairs of proposals containing the most representative information discrepancy for distillation. $\gamma$ controls the proportion of discrepant proposal pairs, further validated in Sec. \ref{ablate}.

\li{For each iteration, we first solve the inner-level optimization, {\em i.e.}, proposal selection, via exhaustive sorting \cite{wu2012maximum}; and then solve the upper-level optimization, distilling the selected pair, based on entropy distillation loss discussed in Sec. \ref{sec3.3}. Considering that there are not too many proposals involved, the process is relatively efficient for the inner-level optimization.} 

\subsection{Entropy distillation loss}
\label{sec3.3}
After selecting a specific number of proposals, we crop the feature based on the proposals we obtained. Most of the SOTA detection models are based on Feature Pyramid Networks (FPN) \cite{lin2017feature}, which can significantly improve the robustness of multi-scale detection. For the Faster-RCNN framework in this paper, we resize the proposals and crop the features from each stage of the neck feature maps. We generate the proposals from the regression layer of the SSD framework and crop the features from the feature map of maximum spatial size. Then we formulate the entropy distillation process as
\begin{equation}
 \mathop{\rm{max}}_{R_n^s}\;p({R_n^s}|{R}_n^t).
 \label{8}
\end{equation}
Here is \textbf{the upper level} of the bi-level optimization, where $\bm{m}$ is solved and therefore omitted. 
\li{We rewrite Equ. \ref{8} and further achieve our entropy distillation loss as}
\begin{equation}
 L_P({\bf w}, \bm{\alpha}; {\gamma}) =({{R}_n^s}-{R}_n^t) + {\rm Cov}({{R}_n^s}, {{R}_n^t})^{-1}({{R}_n^s}-{R}_n^t)^2 + {\rm log}({\rm Cov}({{R}_n^s}, {{R}_n^t})),
 \label{10}
\end{equation}
where ${\rm Cov}({{R}_n^s}, {{R}_n^t}) = \mathbb{E}({{R}_n^s}{{R}_n^t}) - \mathbb{E}({{R}_n^s})\mathbb{E}({{R}_n^t})$ denotes the covariance matrix. 

Hence, we trained the 1-bit student model end-to-end, the total loss for distilling the student model is defined as
\begin{equation}
\small
L = L_{GT}({\bf w}, \bm{\alpha}) +\lambda L_{P}({\bf w}, \bm{\alpha}; {\gamma}) + \mu L_R({\bf w}, \bm{\alpha}),
\label{13}
\end{equation}
where $L_{GT}$ is the detection loss derived from the ground truth label and $L_R$ is defined in Equ. \ref{4}.  

\section{Experiments}
\red{On two mainstream object detection datasets, {\em i.e,} PASCAL VOC \cite{voc2007} and COCO \cite{coco2014}, extensive experiments are undertaken to test our proposed method. First, we go through the implementation specifics of our IDa-Det. Then, in the ablation studies, we set different hyper-parameters and validate the effectiveness of the components as well as the convergence of our method. Moreover, we illustrate the superiority of our IDa-Det by comparing it to previous state-of-the-art 1-bit CNNs and other KD approaches on 1-bit detectors. Finally, we analyze the deploy efficiency of our IDa-Det on hardware.}

\subsection{Datasets and Implementation Details}

\noindent\textbf{PASCAL VOC.}
\red{Natural images from 20 different classes are included in the VOC datasets. We use the VOC {\tt trainval2012} and VOC {\tt trainval2007} sets to train our model, which contain around 16k images, and the VOC {\tt test2007} set to evaluate our IDa-Det, which contains 4952 images. We utilize the mean average precision (mAP) as the evaluation matrices, as suggested by \cite{voc2007}.}

\noindent\textbf{COCO.}
\red{All our experiments on COCO dataset are conducted on the COCO {\tt 2014} \cite{coco2014} object detection track in the training stage, which contains the combination of 80k images and 80 different categories from the COCO {\tt train2014} and 35k images sampled from COCO {\tt val2014}, {\em i.e.}, COCO {\tt trainval35k}. Then we evaluate our method on the remaining 5k images from the COCO {\tt minival}. We list the average precision (AP) for IoUs$\in [0.5:0.05:0.95]$, designated as mAP$@[.5,.95]$, using COCO's standard evaluation metric. For further analyzing our method, we also list AP$_{50}$, AP$_{75}$, AP$_s$, AP$_m$, and AP$_l$.}

\noindent\textbf{Implementation Details.}
\red{Our IDa-Det is trained with two mainstream object detectors, {\em i.e.}, two-stage Faster-RCNN\footnote{In this paper, Faster-RCNN denotes the Faster-RCNN implemented with FPN neck.} \cite{ren2016faster} and one-stage SSD \cite{liu2016ssd}. In Faster-RCNN, we utilize ResNet-18 and ResNet-34 \cite{he2016deep} as backbones. And we utilize VGG-16 \cite{Simonyan15} as the backbone of SSD framework. PyTorch \cite{paszke2017automatic} is used for implementing IDa-Det. We run the experiments on 4 NVIDIA GTX 2080Ti GPUs with 11 GB memory and 128 GB of RAM. We use ImageNet ILSVRC12 to pre-train the backbone of a 1-bit student, following \cite{liu2020reactnet}. The SGD optimizer is utilized and the batch size is set as 24 for SSD and 4 for Faster-RCNN, respectively.}

We keep the shortcut, first layer, and the last layer (the 1$\times$1 convolution layer of RPN and a FC layer of the bbox head) in the detectors as real-valued on Faster-RCNN framework after implementing 1-bit CNNs following \cite{liu2020reactnet}. Following BiDet \cite{wang2020bidet}, the extra layer is likewise retained as real-valued for the SSD framework. Following \cite{wang2020bidet} and \cite{gu2019projection}, we modify the network of ResNet-18/34 with an extra shortcut and PReLU \cite{he2015delving}.

The architecture of VGG-16 is modified with extra residual connections, following \cite{wang2020bidet}. The lateral connection of the FPN \cite{lin2017feature} neck is replaced with 3$\times$3 1-bit convolution for improving performance. 
This adjustment is implemented in 
\begin{wrapfigure}[13]{r}{0.5\textwidth}
	\begin{minipage}[t]{0.5\textwidth}
	\subfigure[Effect of $\mu$.]{
		\begin{minipage}[t]{0.5\textwidth}
			\includegraphics[width=0.95 \linewidth]{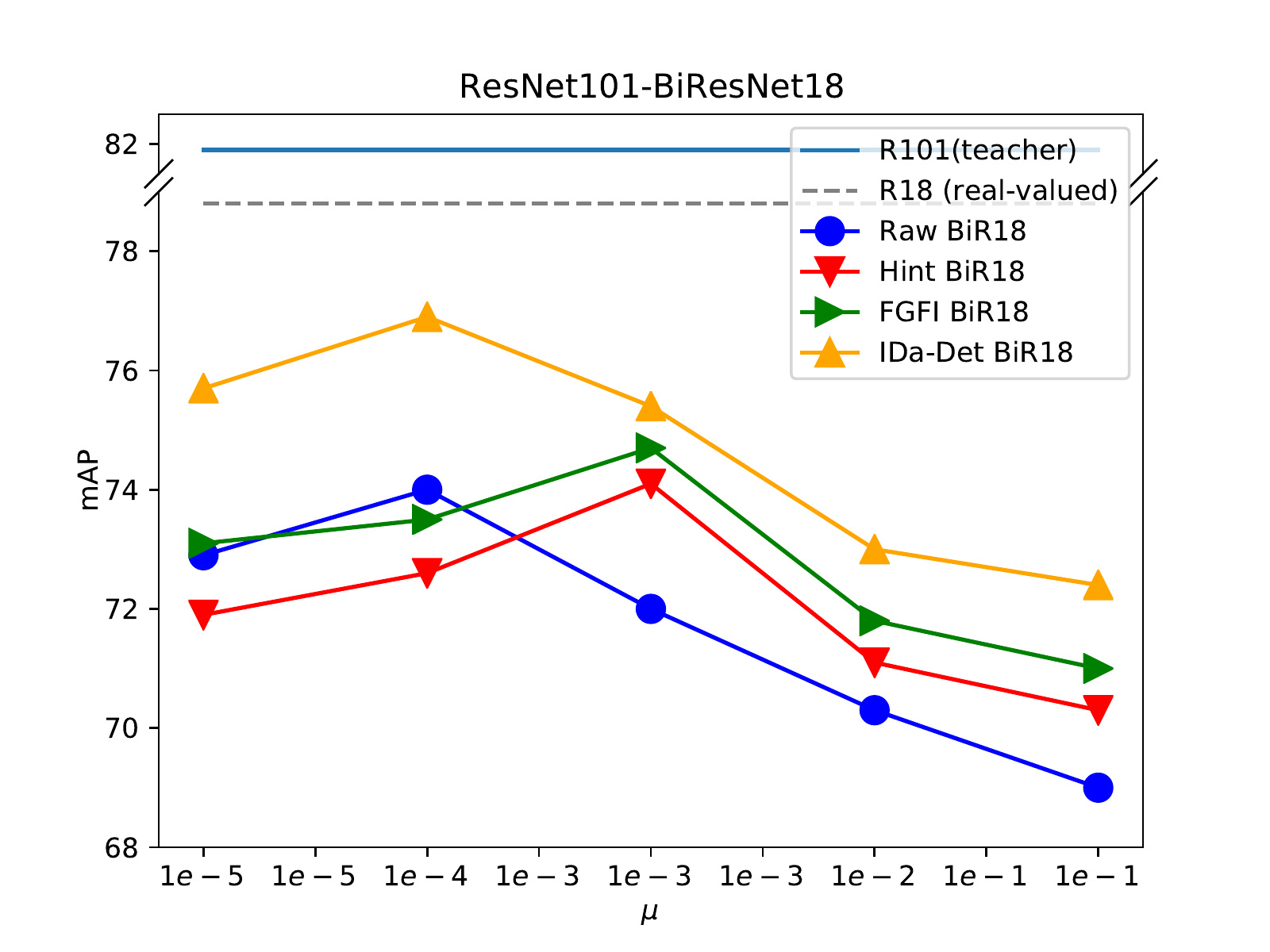}
		\end{minipage}%
	}%
	\subfigure[Effect of $\lambda$ and $\gamma$.]{
		\begin{minipage}[t]{0.5\textwidth}
			\includegraphics[width=0.95 \linewidth]{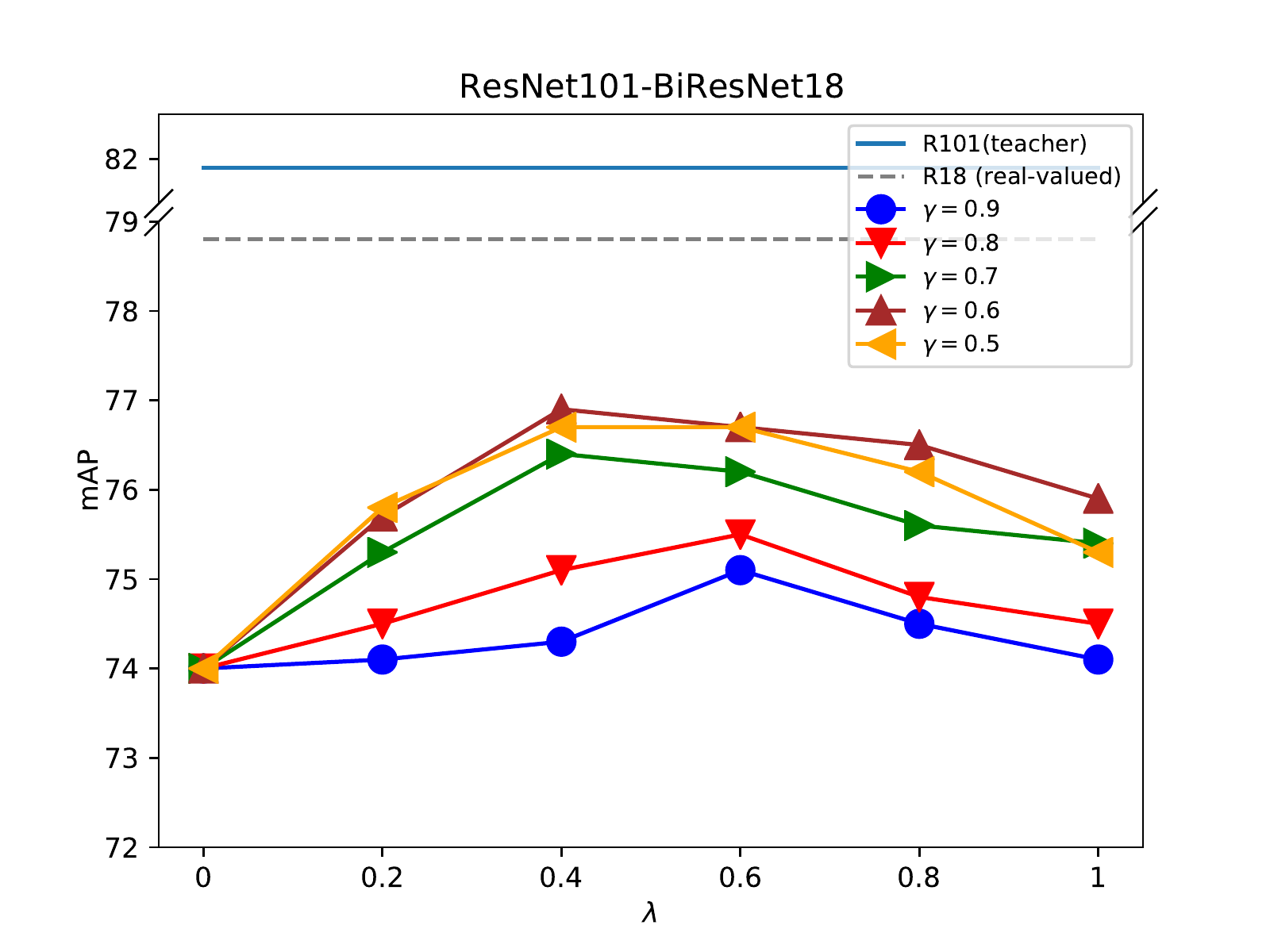}
		\end{minipage}%
	}%
	\caption{On VOC, we (a) select $\mu$ on raw detector and different KD methods including Hint \cite{chen2017learning}, FGFI \cite{wang2019distilling}, and IDa-Det; (b) select $\lambda$ and $\gamma$ on IDa-Det with $\mu$ set as $1e\!\!-\!\!4$.}
	\label{hyper-parameter}
	\end{minipage}
\end{wrapfigure}all of the Faster-RCNN experiments. For Faster-RCNN, we train the model with two stages. 
Only the backbone is binarized at the first stage. Then we binarize all layers in the second stage. Each stage counts 12 epochs. The learning rate is set as 0.004 and decays by multiplying 0.1 in the 9-th and 11-th epochs. We use a loss coefficient set as 5 and multi-scale training method. For fair comparison, all the methods in this paper are implemented with the same training setup. The real-valued counterparts in this paper are  also trained for 24 epochs for fair comparison.
For SSD, the model is trained for 24 epochs with a learning rate of 0.01, which decays to 0.1 at the 16-th and 22-nd epochs by multiplying 0.1.


We select real-valued Faster-RCNN with ResNet101 backbone (81.9\% mAP on VOC and 39.8\% mAP on COCO) and real-valued SSD300 with VGG16 backbone (74.5\% mAP on VOC and 25.0\% mAP on COCO) as teacher network.

\subsection{Ablation Study}
\label{ablate}


\noindent\textbf{Selecting the hyper-parameter.}
As mentioned above, we select hyper-parame\\ters $\lambda$, $\gamma$, and $\mu$ in this part. We first select the $\mu$, which controls the binarization process. As plotted in Fig. \ref{hyper-parameter} (a), we first fine-tune the hyper-parameter $\mu$ controlling the binarization process in four situations: raw BiRes18, and BiRes18 distilled via Hint \cite{chen2017learning}, FGFI \cite{wang2019distilling}, and our IDa-Det, respectively. Overall, the performances increase first and then decrease when increasing the value of $\mu$. On raw BiRes18 and IDa-Det BiRes18, the 1-bit student achieves the best performance when $\mu$ is set as 1$e$-4. And $\mu$ valued 1$e$-3 is better for the Hint, and the FGFI distilled 1-bit student. Thus, we set $\mu$ as 1$e$-4 for an extended ablation study. Fig. \ref{hyper-parameter} (b) shows that the performances increase first and then decrease with the increase of $\lambda$ from left to right. In general, the IDa-Det obtains better performances with $\lambda$ set as 0.4 and 0.6. With varying value of $\gamma$, we find $\{\lambda,\gamma\}$ = \{0.4, 0.6\} boost the performance of IDa-Det most, achieving 76.9\% mAP on VOC {\tt test2007}. Based on the ablative study above, we set hyper-parameters $\lambda$, $\gamma$, and $\mu$ as 0.4, 0.6, and 1$e$-4 for the experiments in this paper.

\noindent\textbf{Effectiveness of components.}
We first compare our information discrepancy-aware (IDa) proposal selecting method with other methods to select proposals: Hint \cite{chen2017learning} (using the neck feature without region mask) and FGFI \cite{wang2019distilling}. We show the effectiveness of IDa on two-stage Faster-RCNN in Tab. \ref{component}.
On the Faster-RCNN, the introducing of IDa achieves improvements of the mAP by 2.5\%, 2.4\%, and 1.8\% compared to non-distillation, Hint, and FGFI, under the same student-teacher framework. Then we evaluate the proposed entropy distillation loss against the conventional $\ell_2$ loss, inner-product loss, and cosine similarity loss. As depicted in Tab. \ref{component}, our entropy distillation loss improves the distillation performance by 0.4\%, 0.3\%, and 0.4\% with Hint, FGFI, and IDa method compared with $\ell_2$ loss. Compared with inner-product and cosine similarity loss, entropy loss outperforms them by 2.1\% and 0.5\% in mAP on our framework, which further reflects the effectiveness of our method.
\begin{table}[t]
\centering
\begin{tabular}{c|c|c|c}
\hline
Model          & \begin{tabular}[c]{@{}c@{}}Proposal\\ selection\end{tabular} & \begin{tabular}[c]{@{}c@{}}Distillation\\ method\end{tabular} & mAP  \\ \hline
Res18          & \XSolidBrush                                                                        & \XSolidBrush                                   & 78.6 \\
BiRes18        & \XSolidBrush                                                                        & \XSolidBrush                                   & 74.0 \\ \hline
Res101-BiRes18 & Hint                                                                                               & $\ell_2$                                                      & 74.1 \\
Res101-BiRes18 & Hint                                                                                               & Entropy loss                                                  & 74.5 \\
Res101-BiRes18 & FGFI                                                                                               & $\ell_2$                                                      & 74.7 \\
Res101-BiRes18 & FGFI                                                                                               & Entropy loss                                                  & 75.0 \\ \hline
Res101-BiRes18 & IDa                                                                                                                & Inner-product                                                 & 74.8 \\
Res101-BiRes18 & IDa                                                                                                                & Consine similarity                                            & 76.4 \\
Res101-BiRes18 & IDa                                                                                                                & $\ell_2$                                                      & 76.5 \\ 
\textbf{Res101-BiRes18} & \textbf{IDa}                                                                                                               & \textbf{Entropy loss}                                                  & \textbf{76.9} \\ \hline
\end{tabular}
\caption{The effects of different components in IDa-Det with Faster-RCNN model on PASCAL VOC dataset. Hint \cite{chen2017learning} and FGFI\cite{wang2019distilling} are used to compare with our information discrepancy-aware proposal selection (IDa). IDa and Entropy loss denote main components of the proposed IDa-Det.}
\label{component}
\end{table}

\begin{table*}[h]
\centering
\setlength{\tabcolsep}{.7mm}{\begin{tabular}{c|c|cc|c|c|c|c}
\hline
Framework                     & Backbone                   & \multicolumn{1}{c|}{\begin{tabular}[c]{@{}c@{}}Quantization\\ Method\end{tabular}} & \begin{tabular}[c]{@{}c@{}}KD\\ Method\end{tabular} & W/A                  & \begin{tabular}[c]{@{}c@{}}Memory Usage \\ (MB)\end{tabular} & \begin{tabular}[c]{@{}c@{}}OPs\\ ($\times10^9$)\end{tabular} & mAP           \\ \hline
\multirow{18}{*}{Faster-RCNN} & \multirow{8}{*}{ResNet-18} & \multicolumn{1}{c|}{Real-valued}                                                   & \XSolidBrush                                        & 32/32                & 112.88                                                       & 96.40                                                        & 78.8          \\ \cline{3-8} 
                              &                            & \multicolumn{1}{c|}{DoReFa-Net}                                                    & \XSolidBrush                                        & 4/4                  & 21.59                                                        & 27.15                                                        & 73.3          \\ \cline{3-8} 
                              &                            & \multicolumn{1}{c|}{ReActNet}                                                      & \XSolidBrush                                        & \multirow{6}{*}{1/1} & \multirow{6}{*}{16.61}                                       & \multirow{6}{*}{18.49}                                       & 69.6          \\
                              &                            & \multicolumn{1}{c|}{Ours}                                                          & \XSolidBrush                                        &                      &                                                              &                                                              & 74.0          \\ \cline{3-4} \cline{8-8} 
                              &                            & \multicolumn{2}{c|}{LWS-Det}                                                                                                             &                      &                                                              &                                                              & 73.2          \\ \cdashline{3-4}
                              &                            & \multicolumn{1}{c|}{Ours}                                                          & FGFI                                                &                      &                                                              &                                                              & 74.7          \\
                              &                            & \multicolumn{1}{c|}{Ours}                                                          & DeFeat                                              &                      &                                                              &                                                              & 74.9          \\ \cdashline{3-4}
                              &                            & \multicolumn{2}{c|}{\textbf{IDa-Det}}                                                                                                    &                      &                                                              &                                                              & \textbf{76.9} \\ \cline{2-8} 
                              & \multirow{8}{*}{ResNet-34} & \multicolumn{1}{c|}{Real-valued}                                                   & \XSolidBrush                                        & 32/32                & 145.12                                                       & 118.80                                                       & 80.0          \\ \cline{3-8} 
                              &                            & \multicolumn{1}{c|}{DoReFa-Net}                                                    & \XSolidBrush                                        & 4/4                  & 29.65                                                        & 32.31                                                        & 75.6          \\ \cline{3-8} 
                              &                            & \multicolumn{1}{c|}{ReActNet}                                                      & \XSolidBrush                                        & \multirow{6}{*}{1/1} & \multirow{6}{*}{24.68}                                       & \multirow{6}{*}{21.49}                                       & 72.3          \\
                              &                            & \multicolumn{1}{c|}{Ours}                                                          & \XSolidBrush                                        &                      &                                                              &                                                              & 75.0          \\ \cline{3-4} \cline{8-8} 
                              &                            & \multicolumn{1}{c|}{Ours}                                                          & FGFI                                                &                      &                                                              &                                                              & 75.4          \\
                              &                            & \multicolumn{1}{c|}{Ours}                                                          & DeFeat                                              &                      &                                                              &                                                              & 75.7          \\ \cdashline{3-4}
                              &                            & \multicolumn{2}{c|}{LWS-Det}                                                                                                             &                      &                                                              &                                                              & 75.8          \\ \cdashline{3-4}
                              &                            & \multicolumn{2}{c|}{\textbf{IDa-Det}}                                                                                                    &                      &                                                              &                                                              & \textbf{78.0} \\ \hline
\multirow{8}{*}{SSD}          & \multirow{8}{*}{VGG-16}    & \multicolumn{1}{c|}{Real-valued}                                                   & \XSolidBrush                                        & 32/32                & 105.16                                                       & 31.44                                                        & 74.3          \\ \cline{3-8} 
                              &                            & \multicolumn{1}{c|}{DoReFa-Net}                                                    & \XSolidBrush                                        & 4/4                  & 29.58                                                        & 6.67                                                         & 69.2          \\ \cline{3-8} 
                              &                            & \multicolumn{1}{c|}{ReActNet}                                                      & \XSolidBrush                                        & \multirow{6}{*}{1/1} & \multirow{6}{*}{21.88}                                       & \multirow{6}{*}{2.13}                                        & 68.4          \\
                              &                            & \multicolumn{1}{c|}{Ours}                                                          & \XSolidBrush                                        &                      &                                                              &                                                              & 69.5          \\ \cline{3-4} \cline{8-8} 
                              &                            & \multicolumn{2}{c|}{LWS-Det}                                                                                                             &                      &                                                              &                                                              & 69.5          \\ \cdashline{3-4}
                              &                            & \multicolumn{1}{c|}{Ours}                                                          & FGFI                                                &                      &                                                              &                                                              & 70.0          \\
                              &                            & \multicolumn{1}{c|}{Ours}                                                          & DeFeat                                              &                      &                                                              &                                                              & 71.4          \\ \cdashline{3-4}
                              &                            & \multicolumn{2}{c|}{\textbf{IDa-Det}}                                                                                                             &                      &                                                              &                                                              & \textbf{72.5} \\ \hline
\end{tabular}}
\caption{We report memory usage, FLOPs, and mAP ($\%$) with state-of-the-art 1-bit detectors, other KD methods on VOC {\tt test2007}. The best results are \textbf{bold}.}
\label{VOC}
\end{table*}

\subsection{Results on PASCAL VOC}
\label{sec4.3}
\red{With the same student framework, we compare our IDa-Det with the state-of-the-art 1-bit ReActNet \cite{liu2020reactnet} and other KD methods, such as FGFI \cite{wang2019distilling}, DeFeat \cite{guo2021distilling}, and LWS-Det \cite{xu2021layer}, in the task of object detection with the VOC datasets. The detection results of the multi-bit quantization method DoReFa-Net \cite{zhou2016dorefa} is also reported. We use the input resolution following \cite{xu2021layer}, {\em i.e.} $1000 \times 600$ for Faster-RCNN and $300 \times 300$ for SSD. }

\red{Tab. \ref{VOC} lists the comparison of several quantization approaches and detection frameworks in terms of computing complexity, storage cost, and the mAP. Our IDa-Det significantly accelerates computation and reduces storage requirements for various detectors. We follow XNOR-Net \cite{rastegari2016xnor} to calculate memory usage, which is estimated by adding 32$\times$ the number of full-precision kernels and 1$\times$ the number of binary kernels in the networks. The number of float operations (FLOPs) is calculated in the same way as Bi-Real-Net \cite{liu2020bi}. The current CPUs can handle both bit-wise XNOR and bit-count operations in parallel. The number of real-valued FLOPs plus $\frac{1}{64}$ of the number of 1-bit multiplications equals the OPs following \cite{liu2020bi}. }

\begin{figure}[t]
\centering
    \begin{minipage}[t]{\linewidth}
	\subfigure[False positives]{
		\begin{minipage}[t]{0.68\textwidth}
			\centering
			\includegraphics[width= \linewidth]{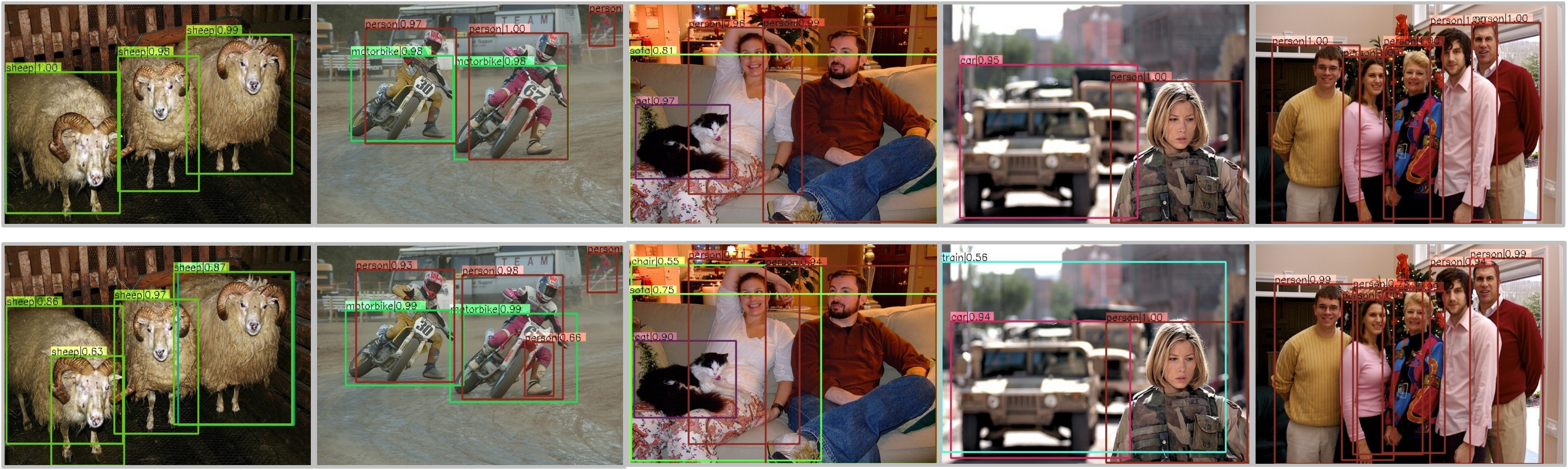}
		\end{minipage}%
	}%
	\subfigure[Missed detection]{
		\begin{minipage}[t]{0.28\textwidth}
			\centering
			\includegraphics[width= \linewidth]{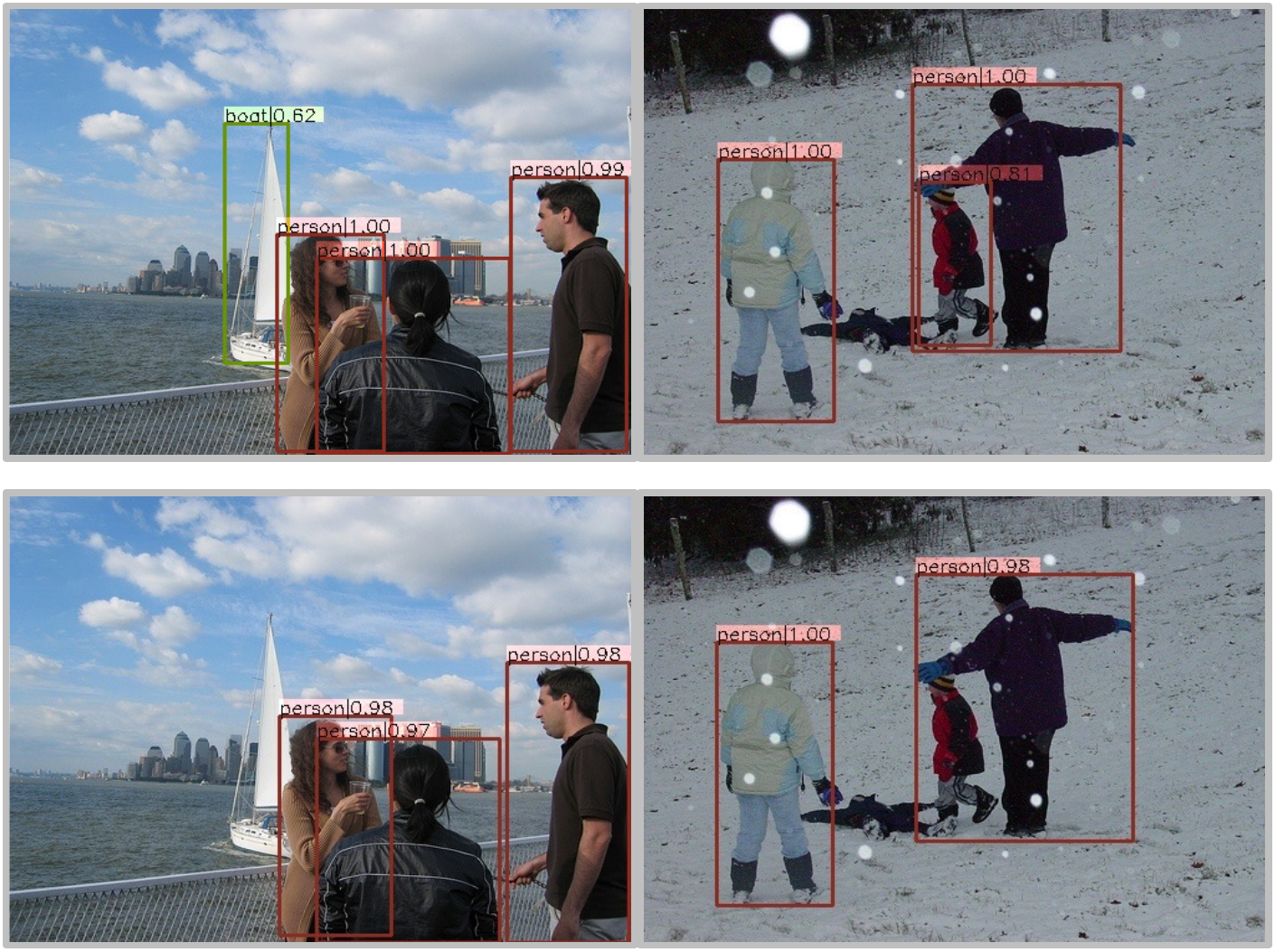}
		\end{minipage}%
	}%
	\caption{Qualitative results on the gain from information discrepancy-aware distilling. The top row shows IDa-Det student’s output. The bottom row images are raw student model’s output without information discrepancy-aware distilling.}
	\label{results}
	\end{minipage}
\end{figure}

\noindent\textbf{Faster-RCNN.}
We summarize the experimental results on VOC {\tt test2007} of 1-bit Faster-RCNNs from lines 2 to 17 in Tab. \ref{VOC}. Compared with raw ReActNet \cite{liu2020reactnet}, our baseline binarization method achieves 4.4\%, and 2.7\% mAP improvement with ResNet-18/34 backbone respectively. 
Compared with other KD methods on the 1-bit detector with the same train and test settings, our IDa-Det surpasses FGFI and DeFeat distillation method in a clear margin of 2.2\%, 2.0\% with ResNet-18 backbone, and 2.6\%, 2.3\% with ResNet-34 backbone. Our IDa-Det significantly surpasses the prior state-of-the-art, LWS-Det, by 3.7\% in ResNet-18 backbone, and 2.2\% in ResNet-34 backbone with the same FLOPs and memory usage. All of the improvements have impacts on object detection.

Compared with the raw real-valued detectors, the proposed IDa-Det surpasses real-valued Faster-RCNN with ResNet-18/34 backbone (\{76.9\%, 78.0\%\} {\em vs.}\{76.4\%, 77.8\%\}) by obviously computation acceleration and storage savings by 5.21$\times$/5.53$\times$ and 6.80$\times$/5.88$\times$. The above results are of great significance in the real-time inference of object detection.

\begin{table}[t]
\centering

\setlength{\tabcolsep}{.05mm}{\begin{tabular}{c|c|cc|c|c|c|c|c|c|c}
\hline
Framework                     & Backbone                   & \multicolumn{1}{c|}{\begin{tabular}[c]{@{}c@{}}Quantization\\ Method\end{tabular}} & \begin{tabular}[c]{@{}c@{}}KD\\ Method\end{tabular} & W/A                  & \begin{tabular}[c]{@{}c@{}}mAP\\ @$[.5, .95]$\end{tabular} & AP$_{50}$     & AP$_{75}$     & AP$_s$        & AP$_m$        & AP$_1$        \\ \hline
\multirow{16}{*}{Faster-RCNN} & \multirow{8}{*}{ResNet-18} & \multicolumn{1}{c|}{Real-valued}                                                   & \XSolidBrush                                        & 32/32                & 32.2                                                       & 53.8          & 34.0          & 18.0          & 34.7          & 41.9          \\ \cline{3-11} 
                              &                            & \multicolumn{1}{c|}{FQN}                                                           & \XSolidBrush                                        & 4/4                  & 28.1                                                       & 48.4          & 29.3          & 14.5          & 30.4          & 38.1          \\ \cline{3-11} 
                              &                            & \multicolumn{1}{c|}{ReActNet}                                                      & \XSolidBrush                                        & \multirow{5}{*}{1/1} & 21.1                                                       & 38.5          & 20.5          & 9.7           & 23.5          & 32.1          \\
                              &                            & \multicolumn{1}{c|}{Ours}                                                          & \XSolidBrush                                        &                      & 26.8                                                       & 46.1          & 27.9          & 14.7          & 28.4          & 36.0          \\ \cline{3-4} \cline{6-11} 
                              &                            & \multicolumn{2}{c|}{LWS-Det}                                                                                                             &                      & 26.9                                                       & 44.9          & 27.7          & 12.9          & 28.7          & 38.3          \\ \cdashline{3-4}
                              &                            & \multicolumn{1}{c|}{Ours}                                                          & FGFI                                                &                      & 27.5                                                       & 46.5          & 28.8          & 15.2          & 28.7          & 37.5          \\
                              &                            & \multicolumn{1}{c|}{Ours}                                                          & DeFeat                                              &                      & 27.9                                                       & 46.9          & 29.3          & 15.8          & 29.0          & 38.2          \\ \cdashline{3-4}
                              &                            & \multicolumn{2}{c|}{\textbf{IDa-Det}}                                                                                                             &                      & \textbf{29.3}                                              & \textbf{48.7} & \textbf{30.9} & \textbf{16.7} & \textbf{29.8} & \textbf{39.9} \\ \cline{2-11} 
                              & \multirow{8}{*}{ResNet-34} & \multicolumn{1}{c|}{Real-valued}                                                   & \XSolidBrush                                        & 32/32                & 35.8                                                       & 57.6          & 38.4          & 21.1          & 39.0          & 46.1          \\ \cline{3-11} 
                              &                            & \multicolumn{1}{c|}{FQN}                                                           & \XSolidBrush                                        & 4/4                  & 31.8                                                       & 52.9          & 33.9          & 17.6          & 34.4          & 42.2          \\ \cline{3-11} 
                              &                            & \multicolumn{1}{c|}{ReActNet}                                                      & \XSolidBrush                                        & \multirow{6}{*}{1/1} & 23.4                                                       & 43.3          & 24.4          & 10.7          & 25.9          & 35.5          \\
                              &                            & \multicolumn{1}{c|}{Ours}                                                          & \XSolidBrush                                        &                      & 29.0                                                       & 47.7          & 30.9          & 16.6          & 30.5          & 39.0          \\ \cline{3-4} \cline{6-11} 
                              &                            & \multicolumn{1}{c|}{Ours}                                                          & FGFI                                                &                      & 29.4                                                       & 48.4          & 30.3          & 17.1          & 30.7          & 39.7          \\
                              &                            & \multicolumn{1}{c|}{Ours}                                                          & DeFeat                                              &                      & 29.8                                                       & 48.7          & 30.9          & 17.6          & 31.4          & 40.5          \\ \cdashline{3-4} 
                              &                            & \multicolumn{2}{c|}{LWS-Det}                                                                                                             &                      & 29.9                                                       & 49.2          & 30.1          & 15.1          & \textbf{32.1}          & \textbf{40.9}          \\ \cdashline{3-4}
                              &                            & \multicolumn{2}{c|}{\textbf{IDa-Det}}                                                                                                             &                      & \textbf{30.5}                                                       &   \textbf{49.2}            &     \textbf{31.8}          &    \textbf{17.7}           &       31.3        &      40.6         \\ \hline
\multirow{8}{*}{SSD}          & \multirow{8}{*}{VGG-16}    & \multicolumn{1}{c|}{Real-valued}                                                   & \XSolidBrush                                        & 32/32                & 23.2                                                       & 41.2          & 23.4          & 5.3           & 23.2          & 39.6          \\ \cline{3-11} 
                              &                            & \multicolumn{1}{c|}{DoReFa-Net}                                                    & \XSolidBrush                                        & 4/4                  & 19.5                                                       & 35.0          & 19.6          & 5.1           & 20.5          & 32.8          \\ \cline{3-11} 
                              &                            & \multicolumn{1}{c|}{ReActNet}                                                      & \XSolidBrush                                        & \multirow{6}{*}{1/1} & 15.3                                                       & 30.0          & 13.2          & 5.4           & 16.3          & 25.0          \\
                              &                            & \multicolumn{1}{c|}{Ours}                                                          & \XSolidBrush                                        &                      & 17.2                                                       & 31.5          & 16.8          & 3.2           & 18.2          & 31.3          \\ \cline{3-4} \cline{6-11} 
                              &                            & \multicolumn{2}{c|}{LWS-Det}                                                                                                             &                      & 17.1                                                       & 32.9          & 16.1          & \textbf{5.5}  & 17.4          & 26.7          \\ \cdashline{3-4}
                              &                            & \multicolumn{1}{c|}{Ours}                                                          & FGFI                                                &                      & 17.7                                                       & 32.3          & 17.3          & 3.3           & 18.9          & 31.8          \\
                              &                            & \multicolumn{1}{c|}{Ours}                                                          & DeFeat                                              &                      & 18.1                                                       & 32.8          & 17.9          & 3.3           & 19.4          & 32.6          \\ \cdashline{3-4}
                              &                            & \multicolumn{2}{c|}{\textbf{IDa-Det}}                                                                                                             &                      & \textbf{19.4}                                              & \textbf{34.5} & \textbf{19.3} & 3.7           & \textbf{21.1} & \textbf{35.0} \\ \hline
\end{tabular}}
\caption{Comparison  with state-of-the-art 1-bit detectors and KD methods on COCO {\tt minival}. Optimal results are {\bf bold}.}
\label{COCO}
\end{table}

\noindent\textbf{SSD.}
On the SSD300 framework with a VGG-16 backbone, our IDa-Det can accelerate computation and save storage by 14.76$\times$ and 4.81$\times$ faster than real-valued counterparts, respectively, as illustrated in the bottom section of Tab. \ref{VOC}. The drop in the performance is relatively minor (72.5\% \;{\em vs.}\;\;74.3\%). Also, our method surpasses other 1-bit networks and KD methods by a sizable margin. Compared to 1-bit ReActNet, our raw 1-bit model can achieve 1.1\% higher mAP with the same computation. Compared with FGFI, DeFeat, and LWS-Det, our IDa-Det exceeds them by 3.0\%, 2.5\%, and 1.1\%, respectively.

As shown in Fig. \ref{results}, BiRes18 achieved by IDa-Det effectively eliminates the false positives and missed detections compared with raw BiRes18. In summary, we achieve new state-of-the-art performance on PASCAL VOC compared to previous 1-bit detectors and KD methods on various frameworks and backbones. We also achieve competitive results, demonstrating the IDa-Det's superiority. 

\subsection{Results on COCO}
\red{Because of its diversity and scale, the COCO dataset presents a greater challenge in the object detection task, compared with PASCAL VOC. On COCO, we compare our proposed IDa-Det with the state-of-the-art 1-bit ReActNet \cite{liu2020reactnet}, as well as the KD techniques FGFI \cite{wang2019distilling}, DeFeat \cite{guo2021distilling}, and LWS-Det \cite{xu2021layer}. We also report the detection result of the 4-bit quantization method FQN \cite{Li_2019_CVPR} and the DoReFa-Net \cite{zhou2016dorefa} for reference. Following \cite{xu2021layer}, the input images are scaled to $1333\times800$ for Faster-RCNN and $300\times300$ for SSD.}

\red{The mAP, AP with different IoU thresholds, and AP of objects with varying scales are all reported in Tab. \ref{COCO}. Due to the constraints in the width of page, we do not report the FLOPs and memory use in Tab. \ref{COCO}. We conduct experiments on Faster-RCNN and SSD detectors, the results of which are presented in the folllowing two parts.}

\noindent\textbf{Faster-RCNN.}
Comparing to the state-of-the-art 1-bit ReActNet, our baseline binarized model achieves a 5.7\% improvement on mAP$@[.5,.95]$ with the ResNet-18 backbone. Compared to state-of-the-art LWS-Det, FGFI, and DeFeat, our IDa-Det prompts the mAP$@[.5,.95]$ by 2.4\%, 1.8\%, and 1.4\%, respectively. With the ResNet-34 backbone, the proposed IDa-Det surpasses FGFI, DeFeat, and LWS-Det by 1.1\%, 0.7\%, and 0.6\%, respectively. IDa-Det, nevertheless, has substantially reduced FLOPs and memory use.

\noindent\textbf{SSD.}
On the SSD300 framework, our IDa-Det achieves 19.4\% mAP$@[.5,.95]$ with the VGG-16 backbone, surpassing LWS-Det, FGFI, and DeFeat by 2.3\%, 1.7\%, and 1.3\% mAP, respectively.

To summarize, our method outperforms baseline quantization methods and other KD methods in terms of the AP with various IoU thresholds and the AP for objects of varied sizes on COCO, indicating IDa-Det's superiority and generality in many application settings.

\begin{table}[t]
\centering
\begin{tabular}{c|c|c|c|c|c}
\hline
Framework                    & Network                    & Method      & W/A   & Latency (ms) & Acceleration             \\ \hline
\multirow{4}{*}{Faster-RCNN} & \multirow{2}{*}{ResNet-18} & Real-valued & 32/32 & 12043.8      & -                        \\\cline{3-6}
                             &                            & IDa-Det     & 1/1   & 2474.4       & 4.87$\times$ \\ \cline{2-6} 
                             & \multirow{2}{*}{ResNet-34} & Real-valued & 32/32 & 14550.2      & -            \\\cline{3-6}
                             &                            & IDa-Det     & 1/1   & 2971.3       & 4.72$\times$         \\ \hline
\multirow{2}{*}{SSD}         & \multirow{2}{*}{VGG-16}    & Real-valued & 32/32 & 2788.7       & -            \\\cline{3-6}
                             &                            & IDa-Det     & 1/1   & 200.5        & 13.91$\times$        \\ \hline
\end{tabular}
\caption{Comparison of time cost of real-valued and 1-bit models (Faster-RCNN and SSD) on hardware (single thread).}
\label{hardware}
\end{table}
\subsection{Deployment Efficiency}
We implement the 1-bit models achieved by our IDa-Det on ODROID C4, which has a 2.016 GHz 64-bit quad-core ARM Cortex-A55. With evaluating its real speed in practice, the efficiency of our IDa-Det is proved when deployed into real-world mobile devices. 
We use the SIMD instruction SSHL on ARM NEON, for making inference framework BOLT \cite{feng2021bolt} compatible with our IDa-Det. We compare our IDa-Det to the real-valued backbones in Tab. \ref{hardware}. We utilize VOC dataset for testing the model. For Faster-RCNN, the input images were scaled to $1000\times600$ and $300\times300$ for SSD. We can plainly see that IDa-Det's inference speed is substantially faster with the highly efficient BOLT framework. For example, the acceleration rate achieves about 4.7$\times$ on Faster-RCNN, which is slightly lower than the theoretical acceleration rate discussed in Sec. \ref{sec4.3}. Furthermore, IDa-Det achieves 13.91$\times$ acceleration with SSD. All deployment results 
in the object detection are significant on real-world edge devices.

\section{Conclusion}
This paper presents a novel method for training 1-bit detectors with knowledge distillation to minimize the information discrepancy. IDa-Det employs a maximum entropy model to select the proposals with maximum information discrepancy and proposes a novel entropy distillation loss to supervise the information discrepancy. As a result, our IDa-Det significantly boosts the performance of 1-bit detectors. Extensive experiments show that IDa-Det surpasses state-of-the-art 1-bit detectors and other knowledge distillation methods in object detection.

\paragraph{{\rm {\bf Acknowledgement.}}} This work was supported in part by the National Natural Science Foundation of China under Grant 62076016, 92067204, 62141604 and the Shanghai Committee of Science and Technology under Grant No. 21DZ1100100.

\clearpage
\bibliographystyle{splncs04}
\bibliography{egbib}
\end{document}